\documentclass{vgtc}                          

\graphicspath{{figures/}{pictures/}{images/}{./}} 

\usepackage{times}                     

\usepackage{tabu}                      
\usepackage{booktabs}                  
\usepackage{lipsum}                    
\usepackage{mwe}                       

\usepackage{mathptmx}                  
\usepackage{enumitem}
\usepackage{amsfonts}
\usepackage{amssymb}
\usepackage{amsmath}
\usepackage[absolute,overlay]{textpos}

\onlineid{0}

\vgtccategory{Research}

\vgtcinsertpkg

\title{TemporalFlowViz: Parameter-Aware Visual Analytics for Interpreting Scramjet Combustion Evolution}

\author{Yifei Jia$^{1,2,\dagger}$
\and Shiyu Cheng$^{1,\dagger}$
\and Yu Dong$^{1}$
\and Guan Li$^{1,2}$
\and Dong Tian$^{1,2}$
\and Ruixiao Peng$^{1,2}$
\and Xuyi Lu$^{1,2}$
\and Yu Wang$^{2,3}$
\and Wei Yao$^{3,*}$
\and Guihua Shan$^{1,2,4,*}$
}

\teaser{
  \centering
  \includegraphics[width=\linewidth]{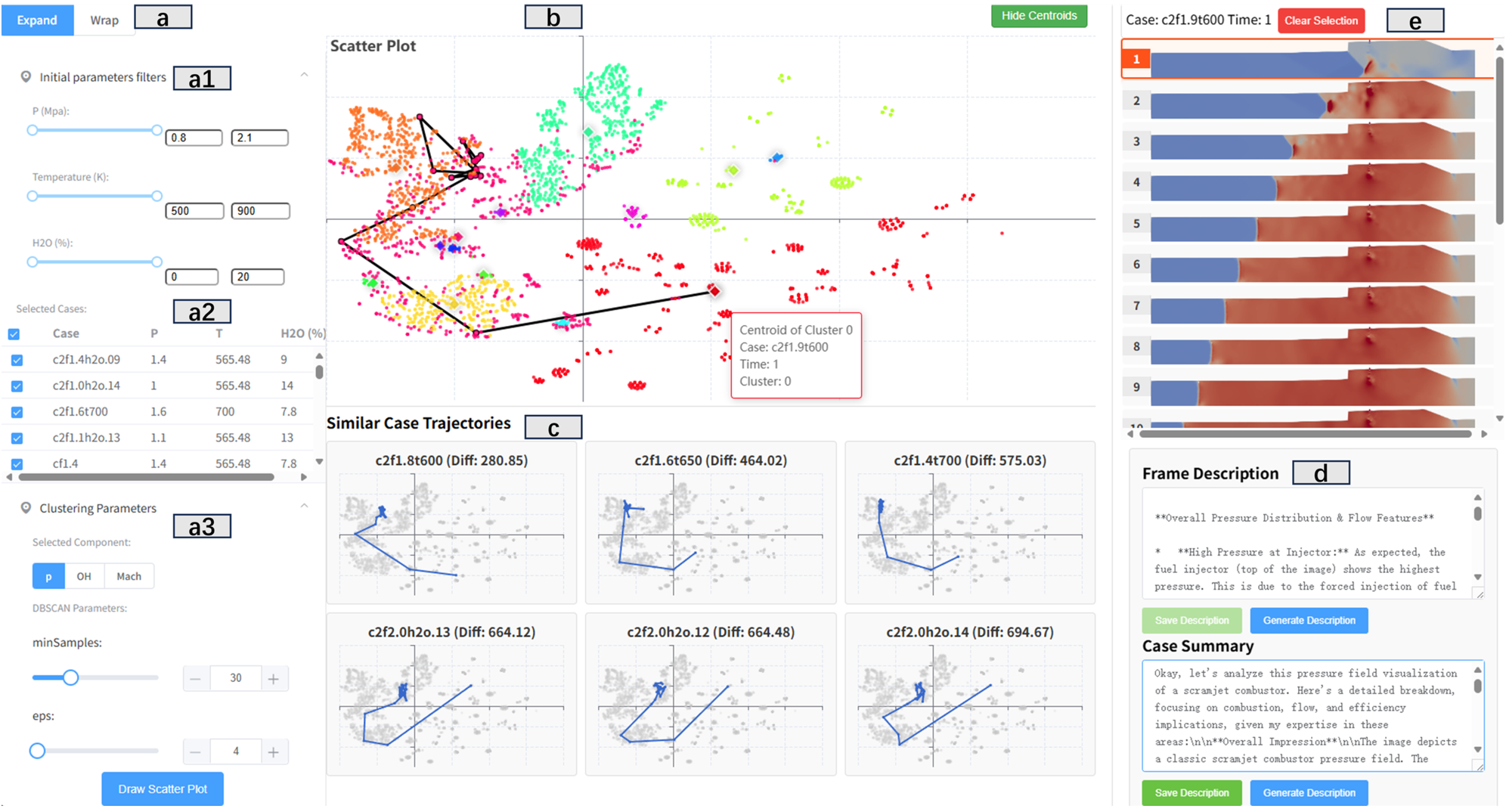}
  \caption{The visual analytics system interface of TemporalFlowViz: (a) the \textit{Filtering Panel}  presents initial parameters filters, case selection table and clustering parameters. (b) the \textit{Temporal Trajectory View} visualizes the clustering result and temporal trajectory of the selected case in latent space. (c) the \textit{Similar Trajectories View }draws the top three most similar temporal trajectories of three simulation cases. (d) the \textit{Report View} shows the AI generated or expert annotated description of selected frame in (b). (e) the \textit{Details View} displays all the flow field frames of the same case of selected point in (b).}
  \label{fig:interface}
}

\abstract{
Understanding the complex combustion dynamics within scramjet engines is critical for advancing high-speed propulsion technologies. However, the large scale and high dimensionality of simulation-generated temporal flow field data present significant challenges for visual interpretation, feature differentiation, and cross-case comparison. In this paper, we present TemporalFlowViz, a parameter-aware visual analytics workflow and system designed to support expert-driven clustering, visualization, and interpretation of temporal flow fields from scramjet combustion simulations. Our approach leverages hundreds of simulated combustion cases with varying initial conditions, each producing time-sequenced flow field images. We use pretrained Vision Transformers to extract high-dimensional embeddings from these frames, apply dimensionality reduction and density-based clustering to uncover latent combustion modes, and construct temporal trajectories in the embedding space to track the evolution of each simulation over time. To bridge the gap between latent representations and expert reasoning, domain specialists annotate representative cluster centroids with descriptive labels. These annotations are used as contextual prompts for a vision–language model, which generates natural-language summaries for individual frames and full simulation cases. The system also supports parameter-based filtering, similarity-based case retrieval, and coordinated multi-view exploration to facilitate in-depth analysis. We demonstrate the effectiveness of TemporalFlowViz through two expert-informed case studies and
expert feedback, showing TemporalFlowViz enhances hypothesis generation, supports interpretable pattern discovery, and enhances knowledge discovery in large-scale scramjet combustion analysis.
} 

\keywords{Visual Analytics, Temporal Flow Field, Scientific Data Simulation, Vision Transformer.}

\begin{document}



\maketitle

\begin{textblock*}{\textwidth}(2cm,24.2cm)  
  \raggedright
  \small
  \textbullet~$^1$Computer Network Information Center, Chinese Academy of Sciences \\
  \textbullet~$^2$University of Chinese Academy of Sciences \\
  \textbullet~$^3$Institute of Mechanics, Chinese Academy of Sciences\\
  \textbullet~$^4$Hangzhou Institute for Advanced Study, UCAS, Hangzhou, 310024, China\\
  \textbullet~$^\dagger$These authors contributed equally to this work.\\
  \textbullet~$^*$Corresponding author: Wei Yao (weiyao@imech.ac.cn), Guihua Shan (sgh@cnic.cn) 
\end{textblock*}

\section{Introduction} 
Supersonic combustion in scramjet engines plays a pivotal role in enabling sustained hypersonic flight. Computational simulations have become indispensable for analyzing the complex interactions among turbulence, chemical kinetics, and thermal effects in these high-speed propulsion systems \cite{annrev1996fluid}. However, the resulting data—typically high-resolution, time-sequenced flow field images across diverse initial conditions—are massive in scale and high in dimensionality, posing significant challenges for downstream analysis and expert interpretation \cite{ttc2009review}.

Traditionally, combustion performance is assessed using scalar metrics such as ignition delay, pressure rise, or thrust efficiency. While useful for summarizing global trends, these metrics offer limited insight into the spatial and temporal structures underlying reactive flows. As domain experts increasingly seek to understand how initial conditions affect combustion dynamics, there is a growing demand for tools that support exploratory, interpretable, and human-in-the-loop analysis of large-scale simulation data.

Recent advances in deep learning offer promising solutions. Vision Transformers (ViTs), in particular, have demonstrated strong capabilities in extracting semantically rich features from images by modeling them as sequences of patches \cite{dosovitskiy2021vit}. Their application to fluid dynamics—such as PDE operator learning \cite{wang2025cvitcontinuousvisiontransformer} and particle image velocimetry \cite{REDDY2025120205}—shows promise for capturing structural complexity in scientific visualizations.

To aid expert interpretation, dimensionality reduction techniques like t-SNE and UMAP \cite{vanDerMaaten2008tsne, mcinnes2018umap}, combined with density-based clustering methods such as DBSCAN \cite{ester1996dbscan}, have been widely used to expose latent patterns, combustion modes, and transitional behaviors in temporal flow field data \cite{Bae2020interactive}. Visual analytics systems further enhance interpretability by linking latent embeddings with raw simulation frames \cite{wu2023vizoptics}, and by enabling multi-view exploration through cluster-based interaction \cite{yu2023user,chen2024geovis}, reasoning\cite{peng2025chinavis}, and trajectory-based interpretation \cite{bostock2011d3, eirich2021irvine,chen2024salientime}.

Despite these advances, key challenges remain. Expert interpretation of flow behavior often relies on implicit knowledge, such as the position of flame fronts, shock structures, or pressure gradients—elements that are not automatically encoded or explained by conventional embedding pipelines. While recent developments in vision–language models (VLMs) offer mechanisms for converting visual input into natural language, their application to scientific visualization is still limited, particularly in domains where domain-specific terminology and interpretive precision are essential.

In this paper, we introduce TemporalFlowViz, a parameter-aware visual analytics workflow and system designed to support expert-driven interpretation of scramjet combustion simulations. Our workflow extracts high-dimensional embeddings from flow field images using pretrained ViTs, applies dimensionality reduction and clustering to identify latent combustion modes, constructs temporal trajectories to trace flow evolution under varying initial conditions, and integrates expert annotations into a VLM-based summarization pipeline to generate descriptive, domain-aligned reports. An interactive multi-view interface connects these components through coordinated filtering, trajectory exploration, case retrieval, and frame-level semantic inspection.

We summarize our contributions as follows:

\begin{itemize}
  \item We propose a unified visual analytics workflow for analyzing temporal flow field data from scramjet combustion simulations. This workflow integrates pretrained Vision Transformer–based embedding, dimensionality reduction and clustering to uncover combustion mode patterns, temporal trajectory modeling to capture flow evolution, and vision–language summarization grounded in expert annotation.

  \item We implement the workflow into an interactive visual analytics system that supports parameter filtering, embedding-space exploration, case similarity retrieval, and coordinated multi-view interpretation of combustion behavior. The system enables experts to compare combustion evolution across cases and interpret dynamic flow structures via both visual and semantic cues.

  \item We conduct two expert-informed case studies and synthesize qualitative feedback to demonstrate the system’s effectiveness. Results show that TemporalFlowViz facilitates hypothesis generation, reveals parameter-sensitive combustion behavior, and enhances the interpretability and communicability of large-scale temporal simulation data.
\end{itemize}

\section{Related Work}

\subsection{Visualization of Combustion Flow Fields}
Visualization techniques for flow fields have evolved to address the challenges of interpreting complex, high‐dimensional fluid dynamics data. Domain‐specific applications such as FlowVisual by Wang et al. have tailored flow visualization for educational contexts, integrating 3D volume rendering and glyph techniques to teach fluid concepts \cite{wang2016flowvisual}. Znamenskaya’s panoramic visualization survey reviews methods for thermophysical flow fields, including refractive index methods and digital particle image velocimetry, emphasizing the need for scalable, quantitative analysis tools \cite{znamenskaya2021methods}. In the era of big data, He et al. presented a parallel visualization framework for large‐scale flow‐field data, combining grid‐segmented data partitioning with optimized communication strategies to achieve real‐time interactivity on millions of grid cells \cite{he2023realtime}. Complementary approaches, such as saliency‐guided importance sampling \cite{Bae2020interactive}, have been proposed to focus visualization effort on regions of greatest dynamical interest.

\subsection{Image Embedding and Latent Pattern Clustering}

\textbf{Image Embeddings.} Effective image representation is fundamental to visual analytics systems, enabling subsequent tasks such as clustering and semantic interpretation. Early handcrafted methods, such as SIFT, extract local keypoint descriptors to capture distinctive image patches invariant to scale and rotation \cite{lowe2004distinctive}. Histogram of Oriented Gradients similarly encodes gradient orientations within localized cells, providing robust structural features for object recognition \cite{dalal2005histograms}. The advent of deep convolutional neural networks marked a paradigm shift: AlexNet demonstrates that end-to-end learning of hierarchical features yields superior performance on large-scale image classification \cite{krizhevsky2012imagenet}, while ResNet introduces residual connections to facilitate training of very deep architectures \cite{he2016deep}. Unsupervised methods, such as stacked denoising autoencoders, learn compact representations by reconstructing corrupted inputs and have been applied successfully to clustering and anomaly detection \cite{vincent2010stacked,Zhao2025,CHEN20241}. More recently, Vision Transformers model images as sequences of patch embeddings, leveraging self-attention to capture long-range dependencies and achieve state-of-the-art results on various benchmarks \cite{dosovitskiy2021vit}. These advances form the foundation for representing complex flow-field imagery in scramjet combustion analysis.

\textbf{Embedding Clustering Methods.} Many studies have been devoted to data clustering methods, and existing methods can be roughly divided into three categories: distance-based methods, density-based methods, and connectivity-based methods. Distance-based methods, such as K-means \cite{6248034, 6817617} and agglomerative clustering \cite{CHIDANANDAGOWDA1978105}, attempt to find relationships between data points based on various distance metrics. Density-based methods attempt to cluster data points by appropriate density functions, including density-based spatial clustering with noise \cite{7340962}. Compared with the previous methods, connectivity-based methods cluster data points into a cluster if they are highly connected. A commonly used method is spectral clustering \cite{ZelnikManor2004SelfTuningSC}. The above ideas form the basis of many methods, such as ensemble clustering \cite{HUANG2016131}, clustering based on non-negative matrix factorization \cite{10.5555/1661445.1661606}, etc.

\subsection{Report Generation on Scientific Data}
Zhang et al. \cite{zhang2025auragenome} introduce an LLM-driven framework for circular genome visualization generation, harnessing LLM agents to automate not only code generation but also text summarization for subsequent analysis. Li et al. \cite{Li_Liang_Hu_Xing_2019} propose a novel approach for generating accurate and robust medical image reports by integrating knowledge-driven encoding, retrieval, and paraphrasing mechanisms. The KERP approach decomposes medical report generation into three main steps: encoding visual features into a structured abnormality graph, retrieving relevant text templates based on detected abnormalities, and paraphrasing these templates to generate the final report. The core of KERP is the Graph Transformer, a novel implementation unit that dynamically transforms high-level semantics between graph-structured data across multiple domains, such as knowledge graphs, images, and sequences. Wang et al. \cite{wang2023chatcad} introduce a framework to integrate LLMs with medical-image CAD networks. The proposed approach uses LLMs to enhance the outputs of multiple CAD networks—including diagnosis, lesion segmentation, and report generation networks—by summarizing and reorganizing information into natural language text. The goal is to combine LLMs’ medical domain knowledge and logical reasoning with the visual understanding capabilities of existing CAD models, creating a more user-friendly and interpretable system for patients compared to traditional CAD systems. R2GenGPT \cite{wang2023r2gengpt} leverages the pre-trained knowledge of LLMs while overcoming their inability to process visual data directly. By aligning image features with the LLM’s text embedding space, the model enables frozen LLMs to generate radiology reports that integrate both visual insights and clinical language, without costly full fine-tuning of the LLM. This approach balances performance and efficiency, making it suitable for real-world medical applications.

\section{Analysis Requirements from Domain Collaboration}
Understanding the dynamic behavior of scramjet combustion processes involves interpreting complex spatial and temporal patterns across numerous simulation runs. To ensure our research addresses real-world analytical challenges, we conducted structured interviews and co-design sessions with two domain experts, E1 and E2. E1 is a researcher specializing in computational fluid dynamics and scramjet combustion modeling, with expertise in reactive flow structures and combustion mode transitions, while E2 is an aerospace engineer engaged in propulsion system development, focusing on simulation-based evaluation of engine configurations and their sensitivity to input parameters of initial conditions.

These two experts represent complementary perspectives from academic and engineering domains. We have collaborated with them over the past two years during a collaborative project involving scramjet combustion simulation analysis. Based on recurring discussions and workflow observations, we distilled a set of representative domain challenges that guided the design of our visual analytics workflow and system.

\subsection{Domain Challenges}

\textbf{C1: High-dimensional temporal data overload.}  
Each simulation of combustion produces dozens of time-sequenced flow field frames. Traditional scalar metrics (e.g., thrust, equivalence ratio) offer limited insights into the spatial and temporal transitions during combustion. Experts expressed the necessity to visually cluster and summarize combustion modes across time.

\textbf{C2: Difficulty in interpreting combustion evolution.}  
Experts have to frequently assess whether combustion reaches steady state, exhibits oscillatory or transitional behaviors, or fails to ignite. However, manual inspection is labor-intensive and susceptible to subjectivity.

\textbf{C3: Limited semantic support for knowledge communication.}  
The interpretation of combustion modes often relies on tacit expert knowledge. There is a lack of structured mechanisms to capture and share this expertise across teams or with automated systems.

\textbf{C4: Complexity in parameter-driven case comparison.}  
Design exploration requires understanding how changes in initial condition parameters affect combustion outcomes. Experts need support for identifying and comparing cases with similar behaviors under different initial conditions.

\subsection{Analytical Requirements}

Based on the above challenges, we derived the following system-level analytical requirements that guided the design of TemporalFlowViz:

\textbf{R1: Clustering and summarizing temporal flow field data} to uncover latent combustion modes and reveal evolution patterns across time.

\textbf{R2: Trajectory-based visualization and comparison} to support stable combustion detection and interpretation of transitional combustion behaviors.

\textbf{R3: Semantic annotation and report generation} to bridge expert interpretations with natural language outputs for communication and automation.

\textbf{R4: Parameter-aware case retrieval and analysis} to enable comparative exploration across simulations with different initial conditions.

These domain-informed requirements serve as the foundation for the system pipeline and interaction design described in the following sections.

\section{Data Background and Simulation Parameters}

To support the parameter-aware analysis tasks described in the previous section, we constructed a dataset of scramjet combustion simulations in collaboration with domain experts. The dataset was generated using a specialized computational fluid dynamics (CFD) platform developed for high-speed reactive flow modeling.

The simulation campaign covers over 200 individual scramjet combustion cases, each conducted under a distinct combination of initial conditions. Each simulation case outputs a time series of flow field data recorded at regular time intervals, capturing the evolution of combustion dynamics from ignition to either a stable combustion state or an instability-induced shutdown. The total dataset includes approximately 3 terabytes of flow field data. Each case is defined by a unique combination of the following three initial condition parameters:

\textbf{Static pressure (P)} — ranging from 0.8 MPa to 2.1 MPa, controlling upstream compression and shock strength;

\textbf{Static temperature (T)} — ranging from 565 K to 830 K, influencing chemical reactivity and ignition delay;

\textbf{Water vapor content (H$_2$O)} — ranging from 7.8\% to 14\%, affecting the mixture equivalence ratio and combustion efficiency.

At each timestep, the CFD solver outputs multiple physical fields including static pressure, OH mass fraction, velocity, Mach number, and others. Among these, the \textbf{pressure field} and \textbf{OH field} were identified by domain experts as particularly important for combustion interpretation. The pressure field reflects global flow evolution and the onset of steady or oscillatory combustion states, while the OH field indicates local combustion activity and flame structure. Each physical field's VTK data is then processed by a python program to map into an image under a uniform colormap and cropped out of excessive background regions to generate a clean visual representation of flow field inside engine.

This dataset serves as the empirical foundation for our visual analytics workflow and enables the exploration of parameter-dependent combustion phenomena across a wide range of initial conditions.

\section{TemporalFlowViz Overview}

To meet the analytical requirements identified in collaboration with domain experts, we presented TemporalFlowViz, an end-to-end visual analytics workflow and system designed to support the parameter-aware interpretation of combustion evolution in scramjet simulations. TemporalFlowViz addresses the four core requirements—temporal summarization (R1), trajectory-based pattern recognition (R2), semantic communication (R3), and parameter-aware case comparison (R4)—via deep feature embedding, temporal trajectory modeling, and expert-guided semantic summarization, all coordinated through a unified visual analytics system.

The TemporalFlowViz consists of three main stages, each corresponding to one or more analytical requirements:

\textbf{Flow Field Data Preprocessing and Embedding (supports R1, R4).} Simulation outputs are organized into time-sequenced flow field frames. These frames are encoded into high-dimensional embeddings and filtered based on user-specified initial conditions (e.g., static pressure, temperature, humidity), enabling comparative analysis across selected cases.

\textbf{Temporal Exploration (supports R1, R2).} Frame-level embeddings are reduced and clustered to reveal latent combustion states. Temporal trajectories are constructed for each simulation by connecting the embeddings of frames chronologically in \textit{Temporal Trajectory View}, allowing experts to assess mode transitions, steady-state convergence, or anomalies.

\textbf{Semantic Summarization and Report Generation (supports R3).} Expert-provided annotations on centroids of representative clusters are used to condition a vision-language model, which generates descriptive summaries of individual frames and entire cases. These outputs help contextualize visual observations and facilitate communication.

\begin{figure}[t]
  \centering
  \includegraphics[width=\linewidth]{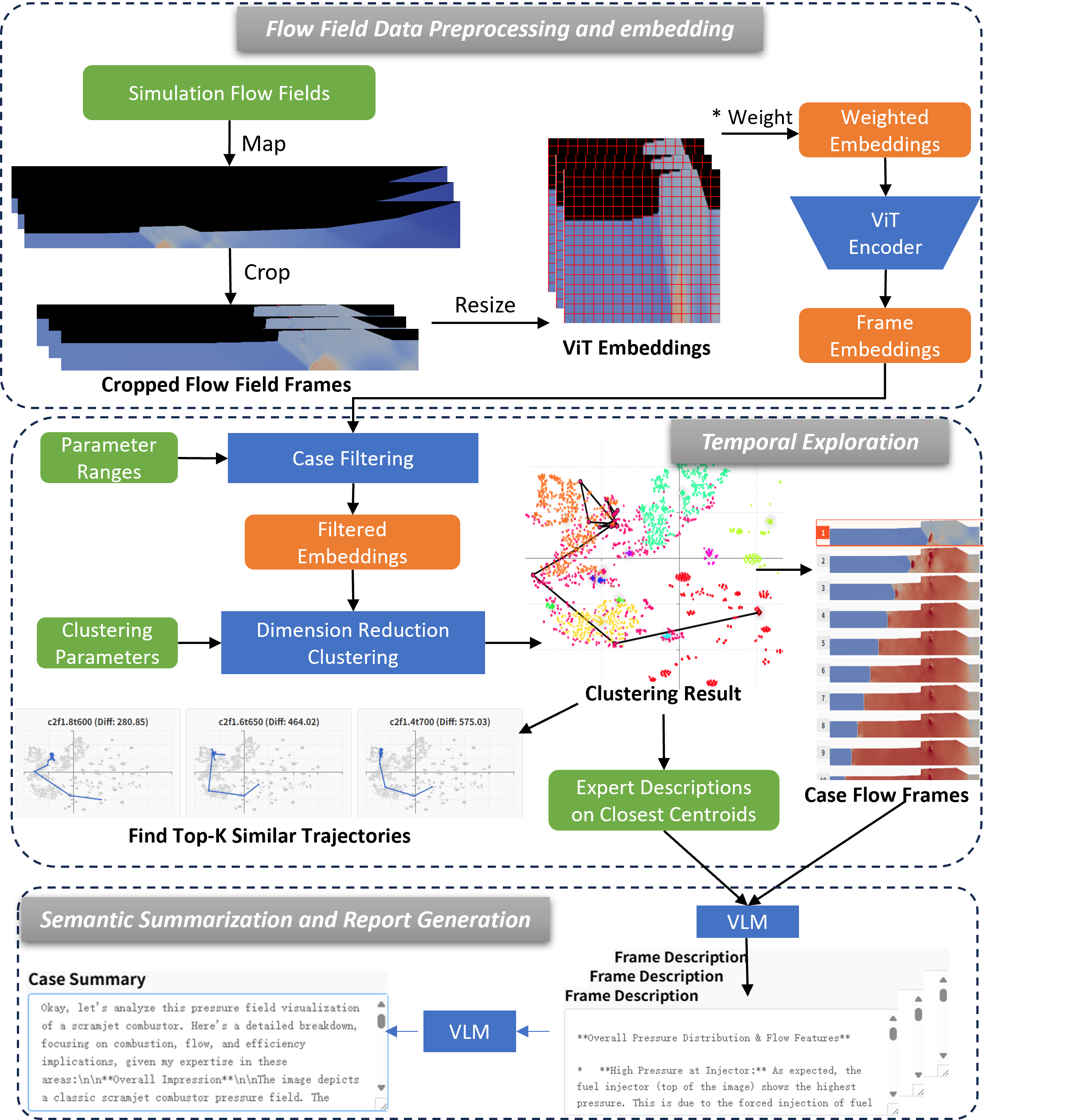}
  \caption{\textbf{Overview of TemporalFlowViz}, including preprocessing and embedding, temporal trajectory-based exploration, and report generation via expert-informed vision-language modeling.}
  \label{fig:pipeline}
\end{figure}

\section{Workflow Design and Experiment}
\label{sec:workflow-design}

We propose a five-stage backend workflow to support the expert-driven visual analytics goals of TemporalFlowViz. The pipeline includes spatially focused preprocessing, feature embedding, dimensionality reduction, temporal trajectory construction, and cluster-based semantic summarization. These components are tightly integrated to translate complex simulation outputs into interpretable latent structures and textual reports.

\subsection{Input Preprocessing and Region Cropping}

TemporalFlowViz begins with preprocessing simulation outputs into multichannel flow field frames, represented such as $X_i(t) = [P_i(t), OH_i(t), \ldots]$, where $P_i(t)$ and $OH_i(t)$ denote the pressure and hydroxyl fields at time step $t$ for simulation case $i$.

In iterative collaboration, both E1 and E2 emphasized that key combustion phenomena—such as ignition onset, shock–flame interaction, and stabilization—primarily manifest near the isolator and cavity zones of the scramjet geometry. Based on this insight, we explored whether focusing on these regions could improve embedding coherence and downstream interpretability.

To test the hypothesis, we compared two preprocessing strategies: using the whole flow field frames versus applying a spatial crop around the isolator and cavity region. We kept the size and location of crop box fixed across all frames of a flow field channel (e.g. pressure field) to ensure comparability. Both inputs were passed through the same embedding extraction pipeline and then projected into the UMAP-reduced latent space.

\begin{figure}[t]
  \centering
  \includegraphics[width=\linewidth]{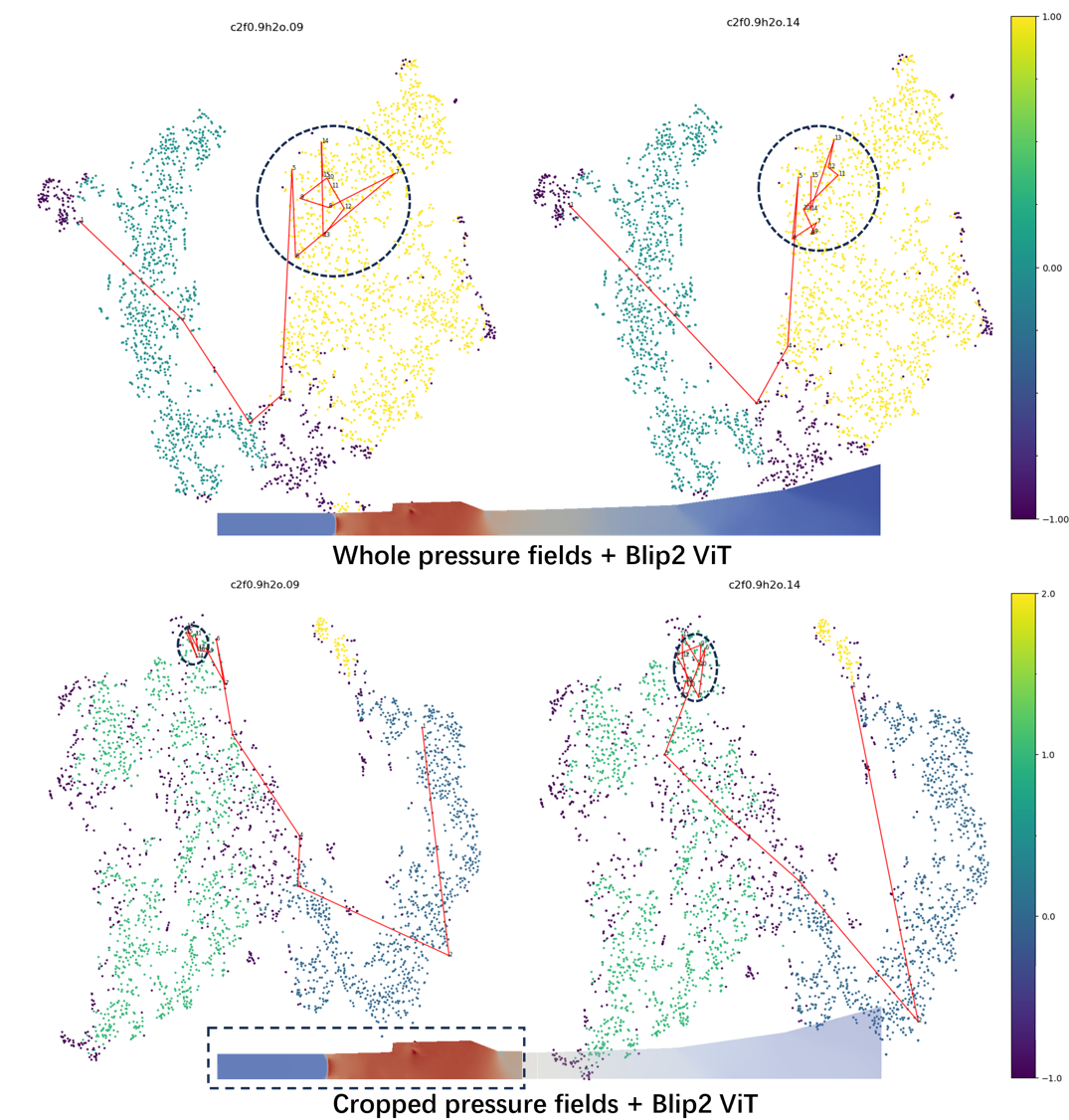}
  \caption{Comparison of UMAP-projected trajectories using pressure field embeddings from whole frames (top 2 graphs) versus cropped regions (bottom 2 graphs). Cropped inputs yield more compact clusters and clearer trajectory convergence.}
  \label{fig:p_crop}
\end{figure}

As shown in \cref{fig:p_crop}, cropped inputs led to tighter clustering and more convergent temporal trajectories, whereas whole-frame inputs produced more scattered and noisy distributions. In our experiment, the average convergence radius r (the average radius of the last $K$ vertices ($K=5$) of all the case trajectories) for PCA reduced embeddings of cropped flow fields ($p_i(t)$), is reduced by $35.24\pm0.59\%$, compared to those of un-cropped flow fields. The convergence radius $r_i$ of a trajectory of case $i$ is defined as: $r_i=\frac{1}{K}\sum_{t=t_i-K+1}^{t_i}{||p_i(t)-\overline{p_i}||}$, where $\overline{p_i}=\frac{1}{K}\sum_{t=t_i-K+1}^{t_i}{p_i(t)}$. These results confirm that spatially focusing on expert-identified regions enhances latent structure clarity and combustion mode separability. Consequently, we adopt cropped inputs centered on the isolator–cavity region as the default configuration throughout our pipeline.

\subsection{Frame Embedding with Vision Transformers}

After preprocessing, each cropped flow field frame $P'_i(t)$ is converted into a high-dimensional embedding vector $p_i(t)$ using a pretrained Vision Transformer (ViT). We define the embedding process as:
\[
p_i(t) = \text{ViT}(P'_i(t), W_P)
\]
where $W_P$ is an optional spatial weighting matrix and $\text{ViT}$ refers to the selected encoder backbone.

To evaluate how different ViT architectures capture combustion-related spatial features, we tested three pretrained models: BLIP2-opt-2.7b \cite{li2023blip2}, InternViT-6B-v2.5  \cite{chen2024internvl}, and AIMv2-1B \cite{fini2024}. For each model, we extracted frame-level embeddings from the same cropped input images and visualized their latent distributions using dimensionality reduction method.

These comparisons revealed that although all models reflect temporal evolution to varying degrees, the latent structures they produce differ significantly in terms of trajectory continuity, cluster separability, and interpretability: BLIP-2 tended to generate smooth, temporally ordered embeddings, with many combustion sequences forming U-shaped manifolds indicative of progressive evolution; InternViT produced multiple fragmented clusters with less clear chronological transitions; AIMv2 resulted in noisier patterns with reduced cluster coherence.

Rather than selecting a single default model, we retain all three ViT backbones as viable options within our system. This allows researchers to experiment with different embedding strategies depending on the analysis task. For example, using BLIP-2 for trajectory-based reasoning, or InternViT to highlight mode transitions via clustering boundaries.

By decoupling embedding from downstream components, TemporalFlowViz supports flexible plug-and-play use of visual encoders while maintaining a consistent analysis pipeline.

\subsection{Dimensionality Reduction for Reasoning}

To interpret and compare the high-dimensional embeddings produced by different ViT models, we applied two commonly used nonlinear dimensionality reduction techniques—t-SNE and UMAP—to project frame-level embeddings $p_i(t) \in \mathbb{R}^n$ into a shared two-dimensional latent space $p'_i(t) \in \mathbb{R}^2$ for visualization and clustering.

\begin{figure}
    \centering
    \includegraphics[width=\linewidth]{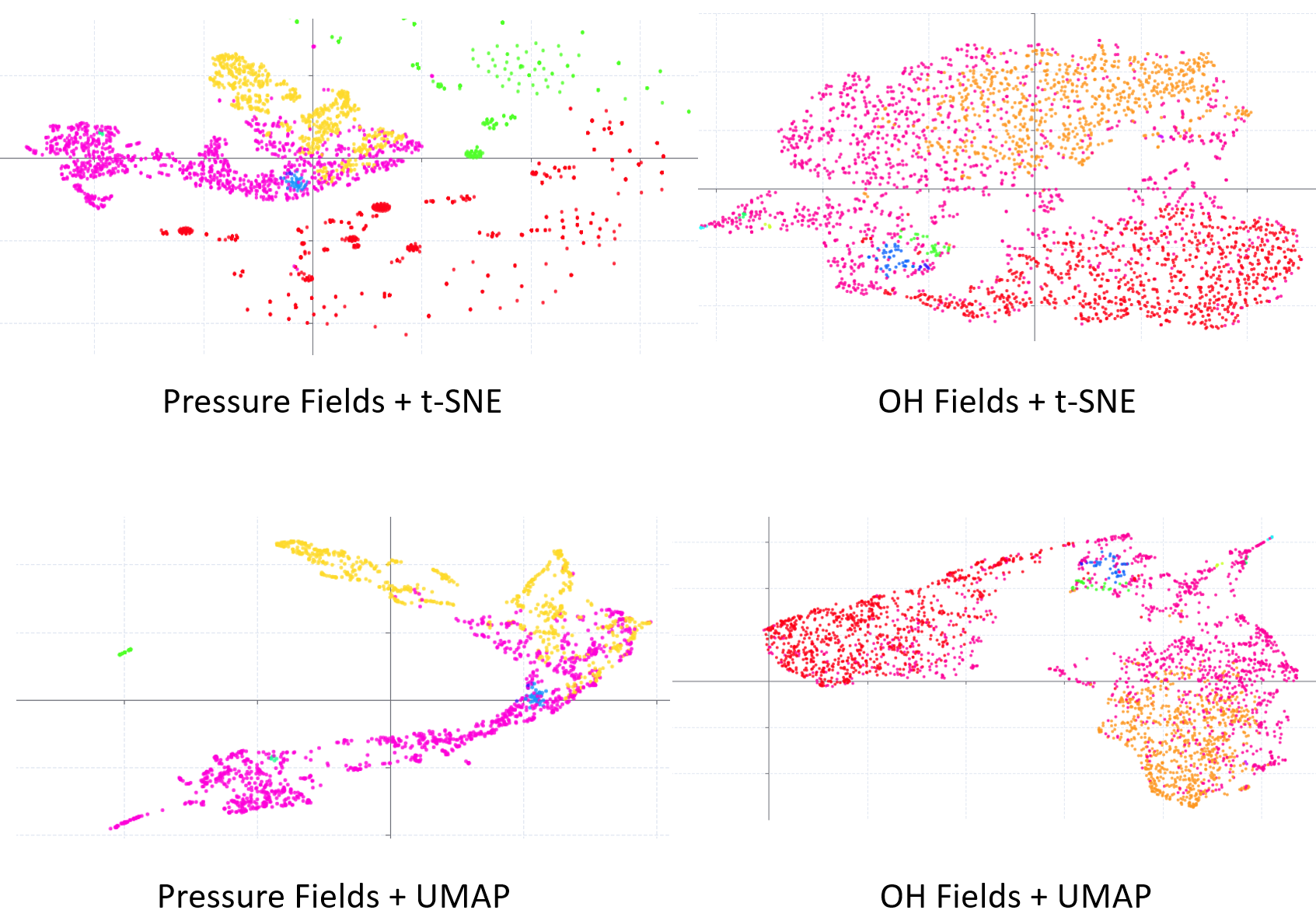}
    \caption{\textbf{UMAP versus t-SNE:} The top two scatter plots display the dimensionality reduction results of the pressure and OH fields' embeddings obtained through t-SNE, while the bottom two plots present the corresponding results using UMAP. UMAP better preserves trajectory shapes and cluster separations at OH fields compared to t-SNE.}
    \label{fig:umapvstsne}
\end{figure}

This projection allows experts to visually assess latent structure, combustion mode separation, and temporal trajectory continuity. As shown in \cref{fig:umapvstsne}, the top two plots illustrate the projection results using t-SNE for both pressure and OH field embeddings, while the bottom two plots show the corresponding UMAP projections.

While t-SNE tends to preserve local neighborhood relationships, it often distorts global layout and trajectory geometry, making it less effective for temporal reasoning. In contrast, UMAP maintains both local and global structural continuity, which better preserves combustion trajectory shapes, cluster boundaries, and transition trends.

Based on visual evaluation and expert feedback, we selected UMAP as the default projection method for OH fields. UMAP not only enables more faithful temporal trajectory construction but also produces cleaner latent clusters that are easier to annotate and interpret.

To ensure comparability across experiments, we fixed the UMAP parameters to $n\_neighbors=15$ and $min\_dist=0.1$, following recommended values from prior work and empirical tuning. This consistent configuration ensures that embeddings from different backbone models can be analyzed under the same projection basis.

\subsection{Trajectory Modeling and Similarity Matching}

To capture the dynamic evolution of combustion behavior over time, we construct latent-space trajectories for each simulation case. Given a case $i$ with $t_i$ time steps, we denote the sequence of UMAP-projected frame embeddings as: $p'_i(1),\ p'_i(2),\ \ldots,\ p'_i(t_i)$. These points are chronologically connected to form a trajectory in the 2D latent space, visually representing the temporal evolution of flow characteristics. Trajectories that gradually converge or oscillate confinedly often indicate stable combustion modes, while diverging ones may suggest unstabilized combustion modes.

To enable comparative reasoning and pattern reuse, we identify the top-$k$ most similar trajectories to any selected case $i$. Let the vertices of trajectory of case $i$ be $p'_i(1), \ldots, p'_i(t_i)$, and let candidate cases be indexed by $j_1, j_2, \ldots, j_k$. The most similar $k$ trajectories retrieved are denoted as $p'_{j_1}, p'_{j_2}, \ldots, p'_{j_k}$.

To quantify trajectory similarity, we have designed and iteratively refined a temporal dissimilarity metric in collaboration with domain experts. Experts evaluated various formulations based on their ability to capture perceptual and physically meaningful similarities—particularly in late-stage behavior related to convergence and stabilization.

The difference between trajectories $p_i(\cdot)$ and $p_j(\cdot)$, revised through multiple rounds of expert review, is defined using a time-warped alignment:

$$
\sum_{(a, b) \in \mathrm{DTW}(t_i, t_j)} \frac{\left\|p_i(a) - p_j(b)\right\|}{\left\|p_i(a) - p_i(a-1)\right\| + \left\|p_j(b) - p_j(b-1)\right\|}
$$

Here, $\mathrm{DTW}(t_i, t_j)$ denotes the optimal alignment path between the two trajectories using Dynamic Time Warping, allowing for flexible temporal matching. $t_i$ and $t_j$ are the number of time steps in cases $i$ and $j$, respectively. This metric maintains normalization by local motion magnitudes while allowing for elastic alignment, thereby addressing minor temporal shifts that do not affect the overall behavior of temporal evolution.

The computed similarity scores serve two purposes. First, they are used to retrieve the top-$k$ most similar cases, which are visualized in the Similar Trajectories View for comparative inspection (see \cref{fig:interface}(c)). Second, they help build a retrieval basis for downstream interpretation, allowing experts to generalize observations from one case to others exhibiting similar dynamics.

\subsection{Semantic Summarization and Report Generation}

To support interpretable summarization of latent flow field structures, we integrate clustering, expert annotation, and vision–language modeling into a unified semantic pipeline.  

We begin by clustering all UMAP-projected frame-level embeddings $p_i(t)$ using DBSCAN. The resulting clusters represent distinct latent combustion modes or transitional stages. For each cluster $C_k$, we identify a representative centroid by selecting the embedding closest to the cluster mean:
\[
\hat{p}_k = \arg\min_{p_i(t) \in C_k} \left\|p_i(t) - \frac{1}{|C_k|} \sum_{p_j(t') \in C_k} p_j(t') \right\|
\]

These centroids form the basis for expert annotation. During workflow execution, domain experts interactively assign combustion-specific terminology to selected centroids through an annotation interface. The resulting annotated exemplars serve as semantic anchors for guiding the generation process.

We compared two open-source VLMs for this task: InternVL2.5-38B and Gemma-3-27B \cite{gemmateam2025gemma3technicalreport}. While InternVL2.5 achieves strong performance on vision-language benchmarks, its deployment presented practical limitations, including high VRAM usage, limited quantization support, and a lack of pre-optimized deployment formats. In contrast, Gemma-3 offers lightweight 4-bit and 8-bit quantizations and efficient image-to-text generation speed. It successfully processes multiple images in a single pass without out-of-memory failures, making it more suitable for our batch-oriented visual analytics workflow. As a result, Gemma-3 is adopted in our workflow deployment.

When selecting a flow field frame for interpretation, the workflow retrieves the top-$k$ nearest cluster centroids in latent space and gathers their associated images and expert-provided annotations. These are used as context inputs to Gemma-3 which generates descriptive summaries of the selected frame. This mechanism supports multiple levels of semantic abstraction. At the frame level, the workflow produces direct semantic interpretations based on nearby cluster exemplars. For entire simulation cases, Gemma-3 aggregates information along latent trajectories to generate case-level overviews. Additionally, when cluster transitions occur—such as during ignition or mode shifts—focused summaries are produced to explain key transitional events like flame stabilization or failure recovery.

\section{Visual Analytics System Design}
\label{sec:system-implementation}

The visual analytics system interface, as illustrated in \cref{fig:interface}, is composed of five tightly coordinated views that correspond to the main stages of the TemporalFlowViz workflow: data filtering, embedding projection, trajectory construction, similarity retrieval, and semantic interpretation. These views support multilevel visual reasoning, enabling experts to move fluidly from parameter-level filtering to latent structure inspection, raw field validation, and case-level reporting.

\subsection{Filtering Panel} 

Filtering Panel (\cref{fig:interface}(a1)) enables users to subset and organize the simulation data based on initial conditions and clustering parameters. The Initial Parameter Filters section provides interactive sliders and text boxes to constrain the simulation dataset by static pressure (P, in MPa), static temperature (T, in Kelvin), and water vapor concentration ($H_2O$, in \%). These filters allow users to isolate specific combustion modes of interest, such as high-pressure lean combustion, and reduce visual clutter in downstream views.

Below the filters, the Selected Cases Table (\cref{fig:interface}(a2)) presents a tabular overview of the filtered simulation cases, showing their initial conditions (parameters) for quick comparison. Users can select multiple cases using checkboxes and initiate visualization by clicking the “Draw Scatter Plot” button. Sorting by column headers allows experts to prioritize cases for analysis.

The Clustering Parameters subsection (\cref{fig:interface}(a3)) allows customization of the DBSCAN clustering algorithm, including the $minSamples$ and $eps$ parameters. Users can also select the flow field component (e.g., pressure, OH concentration, Mach) used to generate feature embeddings. This enables exploration of how different physical variables influence the latent structure and cluster formation.

\subsection{Temporal Trajectory View}

Temporal Trajectory View (\cref{fig:interface}(b)) visualizes high-dimensional frame embeddings projected into a two-dimensional space via dimensionality reduction (e.g., UMAP). Each point represents a frame from a combustion simulation, colored by its cluster assignment. When a case is selected, its frames are chronologically connected to form a temporal trajectory, revealing how combustion dynamics evolve in latent space.

This view supports analysis at multiple levels. Globally, experts can identify clusters corresponding to different combustion modes. Locally, the trajectory of a single case reveals whether the process remains within a stable mode, transitions between modes, or shows oscillatory behavior. Users can zoom, pan, and hover for tooltips. Clicking a point highlights the corresponding frame in the Details View and triggers semantic generation in the Report View. Cluster centroids can be toggled using the ``Hide Centroids / Show Centroids'' button.

\subsection{Similar Trajectories View}

Similar Trajectories View (\cref{fig:interface}(c)) supports comparative reasoning by displaying the top six similar temporal trajectories to the currently selected case. Similarity is computed using a domain-informed difference metric that emphasizes relative deviation near the final time steps of each trajectory, as defined in previous section. This design prioritizes the convergence and stability characteristics that are critical for interpreting combustion behavior.

Each retrieved trajectory is visualized as a mini scatter plot, where the target trajectory is highlighted against a dimmed background of all other points. This layout enables quick visual comparison of trajectory shapes, cluster transitions, and stabilization patterns across cases. Selecting a trajectory in this view updates the main Temporal Trajectory View and allows direct navigation to the corresponding frames in the Details View and Report View, supporting seamless multi-view inspection and cross-case pattern recognition.

\subsection{Details View}

\textit{Details View} (\cref{fig:interface}(e)) provides a scrollable view showing the full temporal sequence of flow field frames of scientific visualizations for the selected case. Each row corresponds to a simulation timestep, presenting the raw field output (e.g., pressure, OH). Users can click a frame to highlight its embedding in the scatter plot and view its description in the Report View.

\textit{Details view} allows experts to validate embedding-based patterns against flow fields, identify key events such as ignition or blowout, and visually confirm the structure of latent trajectories. Hover previews and direct selection support efficient inspection of important frames.

\subsection{Report View}

\textit{Report View} (\cref{fig:interface}(d)) displays the description of the currently selected frame in the scatter plot. Descriptions may be VLM-generated or manually annotated. When generating new frame descriptions, the system retrieves the top-k nearest cluster centroids in latent space and uses their annotated descriptions as context for a vision-language model, which produces a frame-specific description. The images and descriptions of frames from the selected case are used by VLM to generate case summary.

This mechanism links low-level visual data to expert interpretation. By grounding the model in expert-labeled centroids, the generated summaries align with combustion-specific terminology and provide interpretable insights into dynamic phenomena. The Report View supports quick review, edit, and export for documentation or cross-expert discussion.

Together in this interface, these five views form an integrated environment that operationalizes the entire workflow. Through coordinated filtering, embedding visualization, trajectory comparison, raw flow inspection, and semantic interpretation, the system enables experts to explore, interpret, and report on complex combustion simulations with high precision and interactivity.

\section{Case Studies}
To validate the effectiveness of TemporalFlowViz in supporting real-world analytical workflows, we conducted two case studies involving independent exploration. These cases were designed to reflect practical use scenarios and to evaluate how well the system supports expert reasoning, pattern recognition, and semantic interpretation in combustion simulation analysis.

To ensure objective evaluation of the final implementation, two additional experts, E3 and E4, were invited to conduct independent analyses using the deployed system.

Case Study 1 focused on pressure field evolution and combustion mode interpretation and was performed by E3, a postdoctoral researcher specializing in reactive flow diagnostics. Case Study 2 explored the classification of combustion modes based on OH field trajectories and was conducted by E4, a senior simulation engineer engaged in engine configuration evaluation.




\subsection{Case Study 1: Pressure Field Evolution and Combustion Mode Interpretation}

In this case study, we analyzed how pressure field evolution and latent-space trajectories reflect combustion mode transitions, and how the system facilitated interpretation through trajectory-based comparison and automatic report generation.

We first cropped the pressure field images to isolate the core region of the scramjet engine—focusing on the isolator and cavity—and assigned larger weights to subregions with more pronounced combustion activity. The resulting weighted images were encoded into high-dimensional embeddings using Blip2-opt-2.7b, InternViT-6B-v2.5, and AIMv2-1B, and processed through TemporalFlowViz. In all three kinds of ViT-based embeddings tested, the \textit{Temporal Trajectory View} revealed a common structure: the temporal trajectories began in a dispersed region and gradually converged into a compact cluster. To better display the clustering results, we used InternViT to extract embeddings from frames.

\begin{figure}[h]
 \centering 
 \includegraphics[width=\columnwidth]{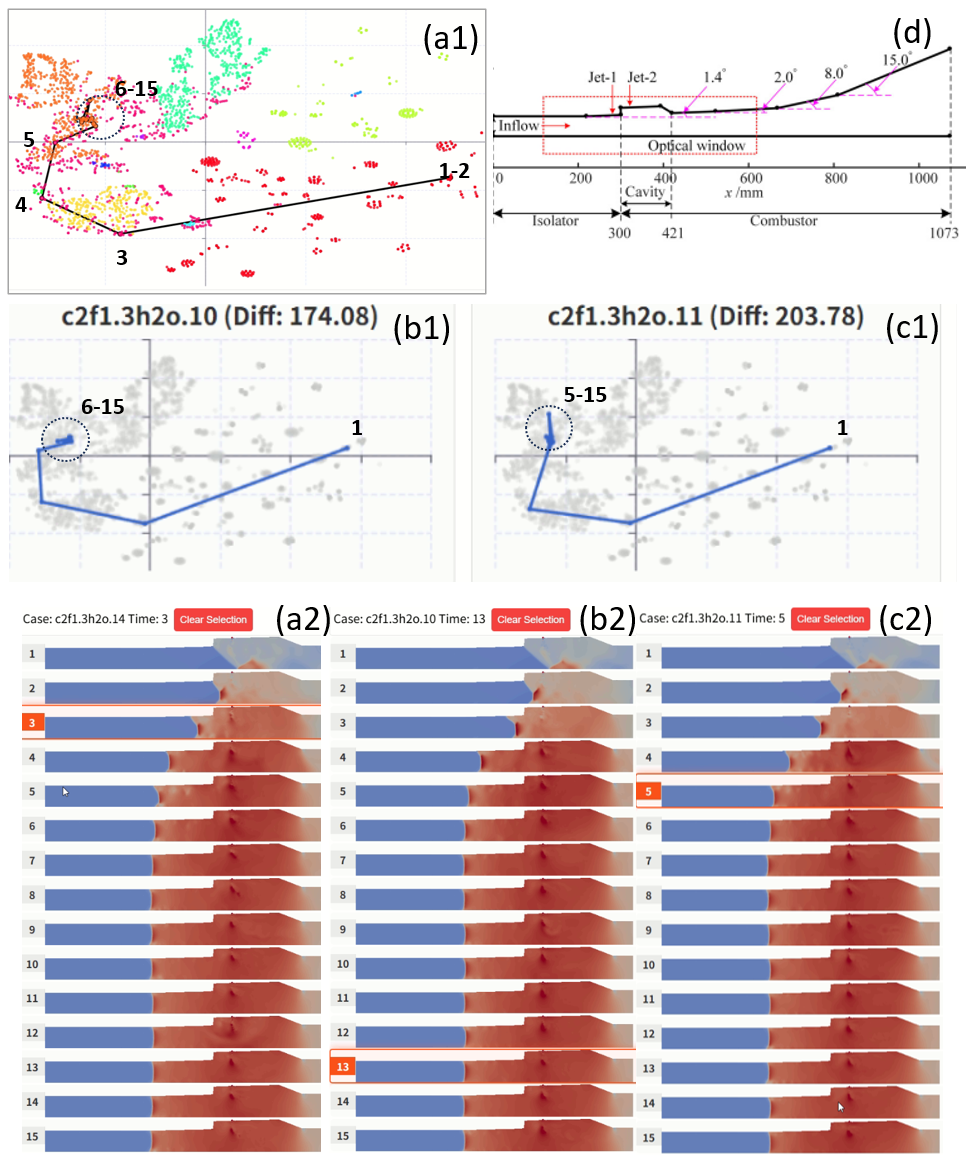}
 \caption{\textbf{The case study 1: Ramjet Pressure field exploration.} Cropped pressure field frames were processed by InternViT and clustered using DBSCAN. Scatter plot (a1) exhibited the clustering result in the \textit{Temporal Trajectory View}. (b1) and (c1) were the 2 most similar trajectories calculated in Similar Trajectory View. (a2), (b2) and (c2) were the flow field frames of the cases in (a1), (b1) and (c1), respectively. (d) was the geometric configurations \cite{shi2021effect} of the scramjet engine in our simulations.}
 \label{fig:case1}
\end{figure}

We filtered simulation cases using the Filtering Panel to constrain initial conditions such as pressure, temperature, and humidity. For three selected cases (a1–c1), we observed the spatiotemporal pressure evolution in \cref{fig:case1}(a2–c2). The blue-red interface indicated a sharp pressure surge. In these cases, the surges appeared well before the cavity—indicative of ramjet-like behavior. As the simulation progressed, the surges either stabilized or oscillated within a narrow region. After 6 ms, the pressure fields remained largely stable in a2, b2, and c2, consistent with steady-state combustion. The first two frames in each series (a2–c2) showed surge positions between the isolator and cavity, a signature of early-stage scramjet ignition.



By comparing the Temporal Trajectory View and the Similar Trajectories View, we confirmed that cases with similar trajectory shapes exhibited consistent pressure field evolution. In particular, cases (a1)–(c1) in \cref{fig:case1} followed highly similar combustion pathways: each began with unstable scramjet ignition and transitioned to stable ramjet combustion by approximately 6 ms. This demonstrated the effectiveness of latent trajectory analysis for understanding dynamic combustion behavior.

\begin{figure}[h]
 \centering 
 \includegraphics[width=\columnwidth]{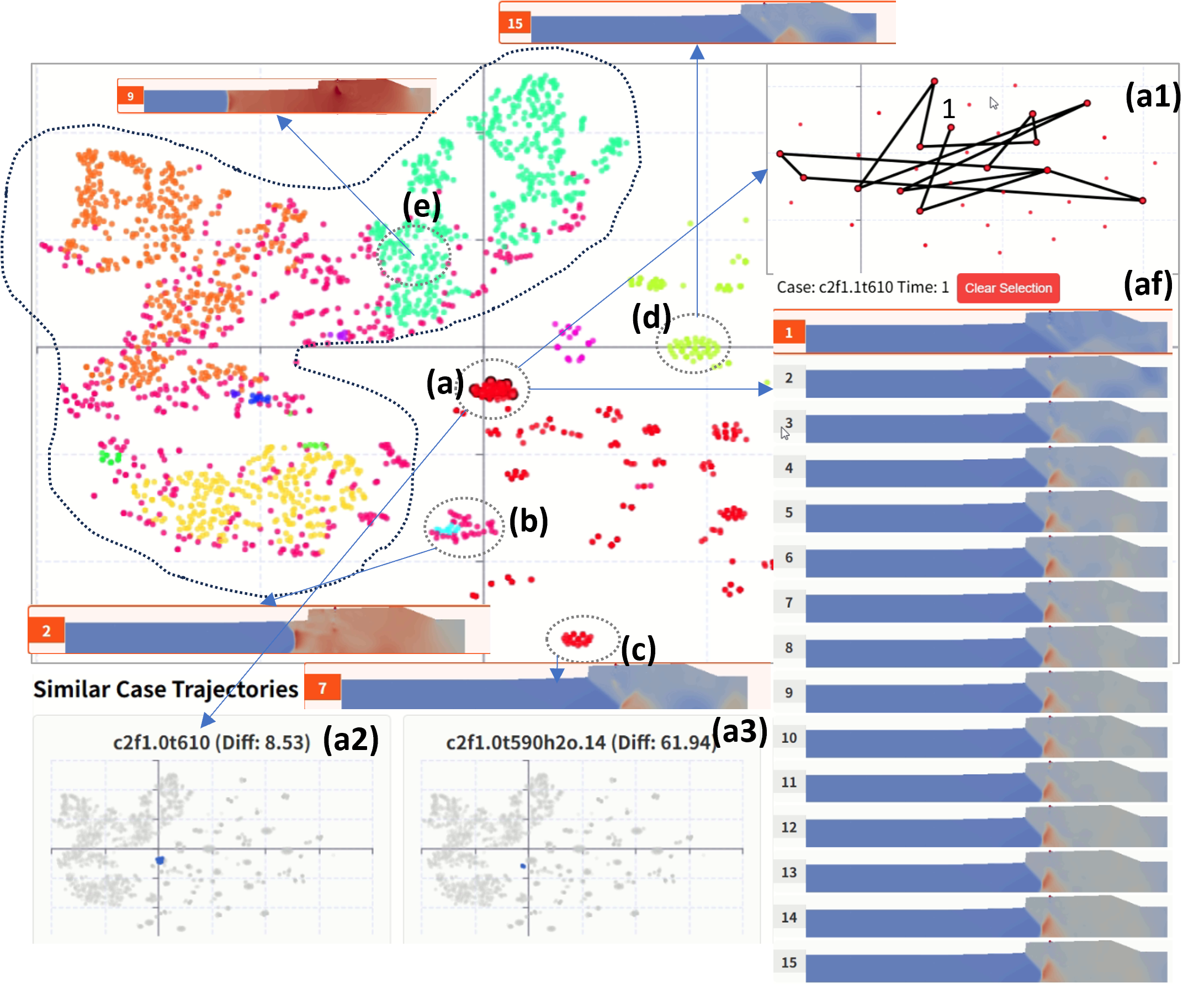}
 \caption{\textbf{The case study 1: Scramjet Pressure fields Exploration.} The temporal trajectory of selected case resided in a small cluster (a), zoomed in to (a1), its flow field frames were in figure (af), and similar trajectories at (a2) and (a3). Clusters (a) to (d) had scramjet combustion mode, while the points in the larger cluster surrounding (e) had ramjet combustion mode.}
 \label{fig:case1s}
\end{figure}

In a separate analysis, we focused on case (a) to study stable scramjet combustion. As shown in \cref{fig:case1s}(a), the trajectory remained within a tight cluster throughout the simulation. A zoomed-in view was shown in \cref{fig:case1s}(a1), while frame snapshots were shown in (af). In this case, the pressure surges consistently occured between the isolator and cavity, and remained stationary—indicative of stable scramjet operation. According to InternViT-based clustering results, trajectories corresponding to stable scramjet combustion always remained within small round clusters, such as circles (a)–(d) in \cref{fig:case1s}. In contrast, ramjet cases typically originated from such clusters and terminated in larger clusters such as (e), which aggregated stable ramjet frames.

\begin{figure}[h]
 \centering 
 \includegraphics[width=\columnwidth]{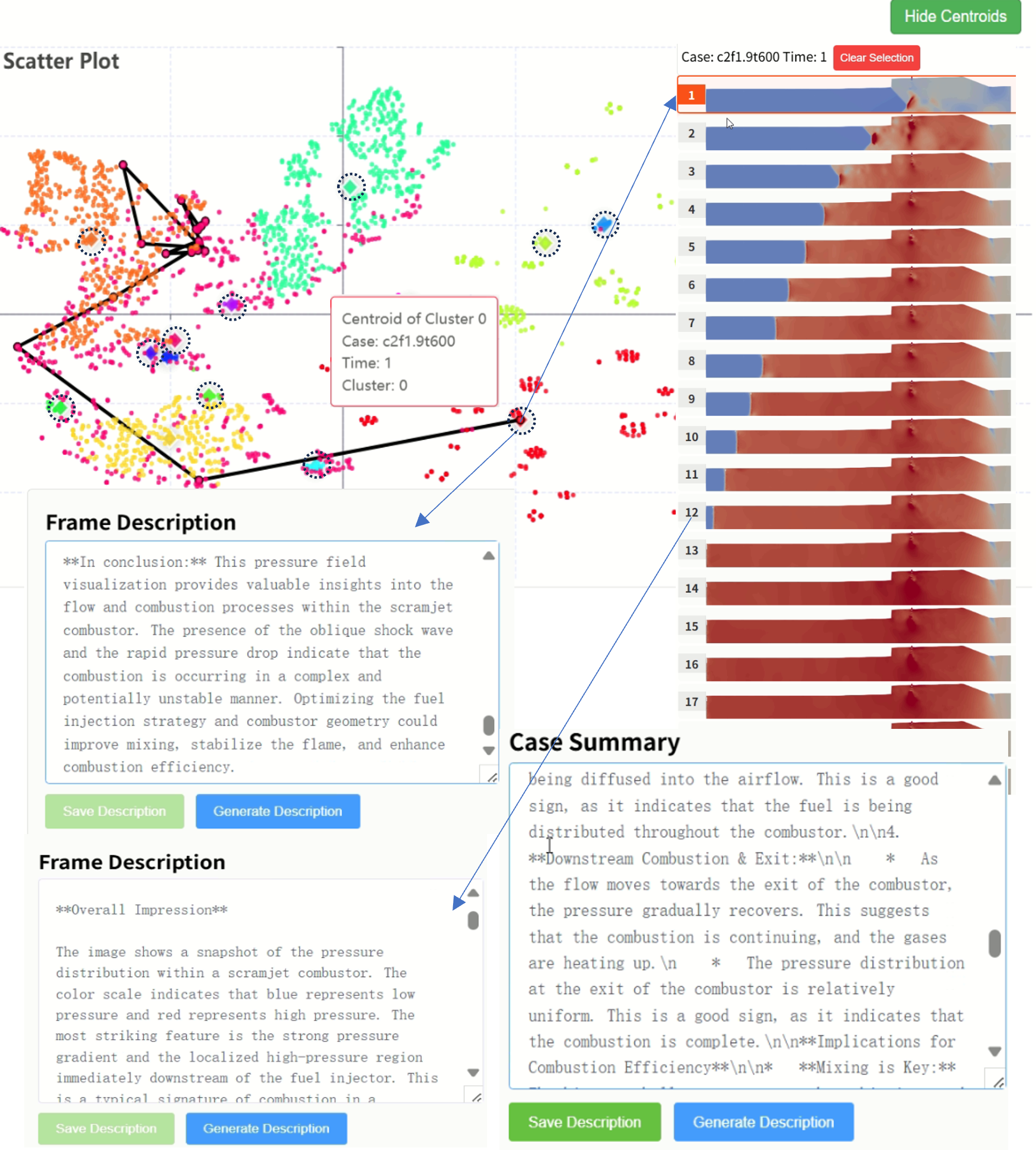}
 \caption{\textbf{The case study 1: Report generation on frames and cases.} Centroids of the clusters were highlighted in diamonds. We drawed dashed circles around them to illustrate. User could inspect or generate frame description on any selected embedding. The description of frames of the selected case were generated from the frame and description of the closest centroids. The case summary was generated from the frames within the case and their descriptions.}
 \label{fig:case1r}
\end{figure}

To evaluate the semantic reporting capability, we collaborated with expert E3 to manually annotate representative cluster centroids. As shown in \cref{fig:case1r}, pressing the \textit{save description} button stored the expert-written description for the centroid of cluster 0 (top-left). The bottom-left panel showed the auto-generated description for the 12th frame of case (a), produced by the Gemma-3 VLM using nearest centroid descriptions and images as context. The right panel showed the case-level summary generated by aggregating frame-level outputs.

Users could access these semantic summaries by clicking on points in the Temporal Trajectory View or selecting frames in the Report View. E3 positively evaluated this feature, citing its ability to accelerate the correction of inappropriate combustion-mode descriptions. The ability to toggle between frame- and trajectory-level summaries allowed for efficient browsing and interpretation of different simulation outcomes.

\subsection{Case Study 2: OH Field Exploration and Combustion Mode Differentiation}

\begin{figure*}[h]
 \centering
 \includegraphics[width=\textwidth]{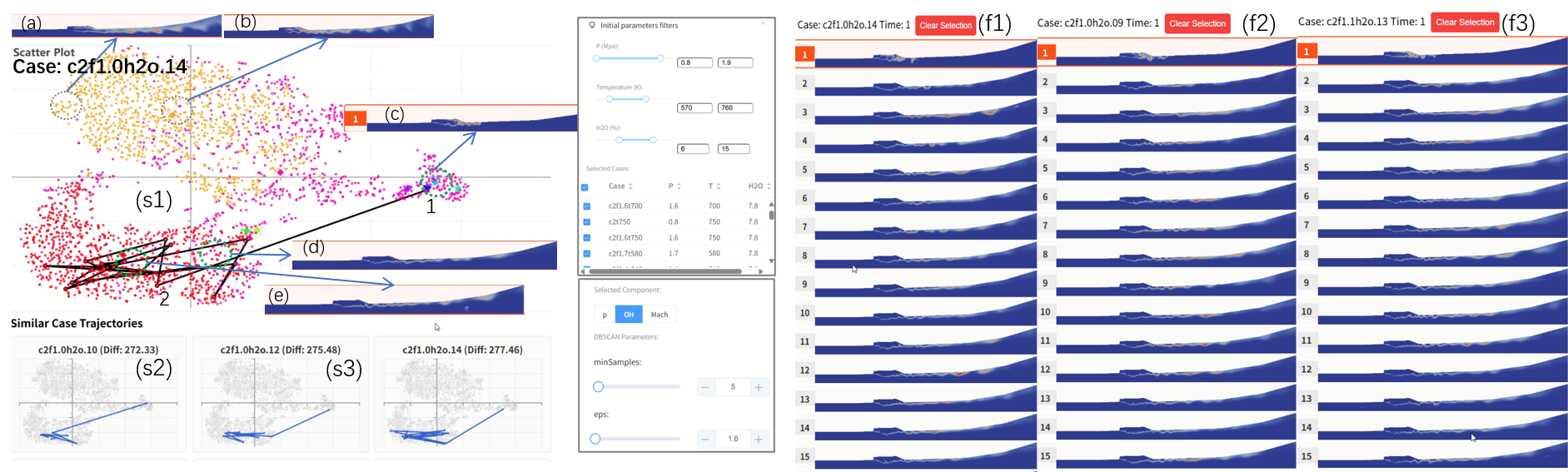}
 \caption{
  \textbf{Case Study 2: Clustering of OH Field Trajectories to Identify Combustion Modes.} The temporal trajectory of the selected case was shown in region (s1), with its two most similar trajectories located at (s2) and (s3). The corresponding OH field images for these cases were displayed in (f1)–(f3). Two distinct combustion modes were identified: the red cluster corresponds to the \emph{shear layer combustion mode}, characterized by narrow, stratified flame structures along the fuel–air interface (see circled regions in (d)–(e)); the yellow cluster corresponded to the \emph{jet-wake combustion mode}, where the flame formed a wide, curved tail extending from the fuel inlet (see circled regions in (a)–(b)). The clustering results demonstrated that the system could effectively separate and characterize combustion behaviors across different cases, even when OH field trajectories fluctuated at steady state.
 }
 \label{fig:case2}
\end{figure*}

In this case, we focused on the hydroxyl radical (OH) field, which served as an important indicator of combustion activity. To reduce noise and enhance spatial interpretability, the OH field images were preprocessed by cropping them to retain only the core scramjet engine region, removing irrelevant background areas. Based on domain knowledge, we assigned higher weights to the upper region of the engine, where the OH field exhibited stronger variation and contains more informative flame structures. Less active regions, such as the left portion of the domain, received smaller weights. These weighted OH field images were then encoded into high-dimensional embeddings using a pretrained BLIP-2 Vision Transformer (ViT) model, capturing localized combustion characteristics across time.

Using the Filtering Panel in visual analytics system, E4 interactively adjusted the ranges of initial conditions, along with DBSCAN clustering parameters including $\varepsilon$ and $minSamples$, to refine the separation of trajectory groups. As illustrated in \cref{fig:case2}, the temporal trajectory of the selected case was shown in region (s1), with its two most similar trajectories located at (s2) and (s3). The corresponding OH field snapshots were presented in regions (f1)–(f3).

The clustering results revealed two distinct combustion modes. The red cluster in the scatter plot corresponded to the \emph{shear layer combustion mode}, characterized by narrow, stratified flame structures forming along the fuel–air interface. This could be observed in the circled regions of \cref{fig:case2}(d)(e), where the OH field intensity was concentrated along a thin horizontal band. In contrast, the yellow cluster corresponded to the \emph{jet-wake combustion mode}, where the flame emerges directly from the fuel inlet and forms a broad, curved tail. This was evident in the circled regions of \cref{fig:case2}(a)(b), where a large flame wake propagated along the upper part of the field.

Interestingly, E4 also observed that even after the pressure field of a simulation reaches a steady state, the OH field trajectory may continue to fluctuate within a relatively wide region of the embedding space. However, these fluctuations consistently remain within the same cluster. This suggested that while the flame shape exhibited spatial variability, the underlying combustion mode remained stable. This behavior reflected the system's ability to distinguish transient structural changes from actual combustion mode transitions—enabling robust combustion classification despite internal oscillations.

Overall, expert E4 highlighted that the combined use of trajectory views, spatial clustering, and multiframe comparison enabled rapid identification of representative combustion modes and supported diagnostic insight across multiple simulation cases.

\section{Discussion}

\subsection{Expert Feedback and Insights}
In addition to the structured analyses presented in our case studies, we also conducted post-session interviews with the domain experts to gather broader impressions of system usability, interpretability, and analytical value. These sessions aimed to evaluate not only how well the system supported specific tasks, but also how experts perceived its potential for generalization to larger workflows and collaboration scenarios. The following summarizes key insights drawn from these expert reflections.

First, experts consistently emphasized the benefit of visual-semantic alignment enabled by our workflow. By allowing cluster centroids to be annotated and reused across simulations, the system supports knowledge transfer and interpretive consistency across teams. One expert highlighted that “labeling once and seeing it echoed across dozens of similar cases” allowed them to validate combustion behavior at scale without redoing frame-by-frame inspections.

Second, the semantic report generation was found to be valuable not only as an interpretive tool, but also as a mechanism for cross-expert communication. The VLM-generated summaries, grounded in human-provided insights, served as a shared semantic layer that all experts could use to explain, compare, or question simulation behavior—particularly in unfamiliar parameter regimes. This capacity to produce interpretable narratives from latent structures significantly lowers the barrier to entry for less experienced analysts or interdisciplinary collaborators.


Third, the interactive views—particularly the combination of the \textit{Temporal Trajectory View} and the \textit{Similar Trajectories View}—were praised for facilitating rapid understanding and intuitive reasoning about the simulation data. Experts could trace how ignition and stabilization evolved in the embedding space, and then explore comparable cases with a single click. Several experts noted that this not only helped them validate existing hypotheses, but also led to the discovery of unexpected cases exhibiting hybrid combustion characteristics.

Finally, experts viewed the system as scalable to high-volume simulation settings, where hundreds of cases must be triaged or grouped. By turning latent trajectory structure into interpretable visual clusters and linking them to semantic summaries, the system allows domain specialists to move from raw images to hypothesis-driven analysis at a significantly accelerated pace.

\subsection{Limitations and Opportunities}

While TemporalFlowViz demonstrates strong performance and expert usability across combustion interpretation tasks, several design choices present opportunities for future enhancement. Currently, clustering relies on manually selected parameters (e.g., $\varepsilon$, $minSamples$), which allows fine control but may benefit from adaptive tuning strategies for improved consistency across variable datasets.

The pretrained Vision Transformers used in our pipeline are sourced from general-purpose corpora. Despite their success in capturing combustion-relevant features in pressure and OH fields, task-specific fine-tuning could further enhance sensitivity to subtle flow structures and transitional events.

Expert annotation of cluster centroids has proven effective in anchoring semantic generation, but it still introduces subjectivity. To scale this process, future versions may incorporate structured annotation templates, provenance tracking, or collaborative validation mechanisms to streamline consensus building.

Although our case studies focused on steady and quasi-steady combustion, the system is designed to generalize across unsteady or transient events. Preliminary tests indicate that the current trajectory modeling can serve as a foundation for more expressive representations—such as hierarchical time-scale embedding or event-aware trajectory segmentation—making TemporalFlowViz adaptable to a wider range of phenomena including blowout or re-ignition.

These limitations suggest directions for refinement rather than structural shortcomings, and do not detract from the system’s core contributions in enabling interpretable, expert-guided visual analysis of combustion simulation data.

\section{Conclusion and Future Work}

In this paper, we presented TemporalFlowViz, a visual analytics workflow and system for expert-guided exploration, clustering, and interpretation of temporal flow-field data from scramjet combustion simulations. TemporalFlowViz combines ViT embeddings, interactive trajectory construction, density-based clustering, and expert-augmented semantic summarization to support interpretable analysis of high-dimensional, time-sequenced combustion data. By allowing experts to annotate representative cluster centroids, which then condition a vision–language model for generating frame- and case-level descriptions, we establish a reusable semantic layer that transforms complex flow field patterns into interpretable textual narratives. This multimodal annotation pipeline enables automated reporting, similarity-driven retrieval, and collaborative analysis with minimal manual overhead. We demonstrated the utility of TemporalFlowViz through two expert-driven case studies focusing on pressure and OH field dynamics. These studies revealed the system’s ability to distinguish combustion modes, uncover evolution patterns, and support scalable interpretation across parameterized simulations. Expert feedback confirmed that the system accelerates hypothesis validation, reduces cognitive load, and facilitates knowledge sharing across teams. In summary, TemporalFlowViz represents a step toward deeply integrated, expert-informed visual analytics for scientific simulations—fusing deep learning, trajectory-based reasoning, and semantic interpretability into a unified framework for understanding combustion dynamics at scale.

In future, we identify two main directions for future research. First, we plan to construct a domain-adaptive Vision Transformer through fine-tuning on large-scale combustion simulation datasets. This will enhance the sensitivity of embeddings to subtle physical structures such as ignition fronts and weak flame zones. Second, we aim to develop a multimodal annotation workflow powered by large vision–language models to support scalable, automated labeling of scientific phenomena. This will reduce expert workload and help surface recurring knowledge patterns across large simulation corpora.

\section*{Open Source Availability}
The code and video demonstrated for this paper are available at: \\
\url{https://github.com/JYF98/temporalflowviz}

 \acknowledgments{We thank all reviewers for their constructive comments. This work was supported by the Strategic Priority Research Program of Chinese Academy of Sciences (Grant No. XDB0500103) and the National Natural Science Foundation of China (No. 62202446). }


\bibliographystyle{abbrv-doi}

\bibliography{jov.bbl}
\end{document}